\documentclass[letterpaper, 10 pt, conference]{ieeeconf}  % Comment this line out if you need a4paper

\IEEEoverridecommandlockouts                              % This command is only needed if 
\usepackage[T1]{fontenc}
\usepackage[utf8]{inputenc}

\usepackage{amsmath,amssymb,amsfonts}
\usepackage{algorithmic}
\usepackage{graphicx}
\usepackage{textcomp}
\usepackage{xcolor}
\usepackage{stfloats}
\usepackage{multirow}
\usepackage{booktabs}
\usepackage{siunitx}
\usepackage{makecell}
\usepackage{caption}
\usepackage{hyperref}
\usepackage{hhline}

\usepackage{graphicx}
\usepackage{mwe}   %%%% PW added for the image

\newcommand{\norm}[1]{\left \lVert #1 \right \rVert}

\title{\LARGE \bf
Latent Conditioned Loco-Manipulation Using Motion Priors
}

\author{Maciej Stępień$^{1, 2}$, Rafael Kourdis$^{1, 3}$, Constant Roux$^{1}$ and Olivier Stasse$^{1, 2}$% <-this % stops a space
\thanks{$^{1}$LAAS-CNRS, Université de Toulouse, France, {\tt\small firstname.lastname@laas.fr}}%
\thanks{$^{2}$Artificial and Natural Intelligence Toulouse Institute, France}
\thanks{$^{3}$University of Edinburgh, School of Informatics, UK}%
}

\begin{document}

% including picture at the top from https://arxiv.org/abs/2403.18765
\makeatletter
\let\@oldmaketitle\@maketitle% Store \@maketitle
\renewcommand{\@maketitle}{\@oldmaketitle% Update \@maketitle to insert...
  \centering \includegraphics[width=1.0\linewidth]{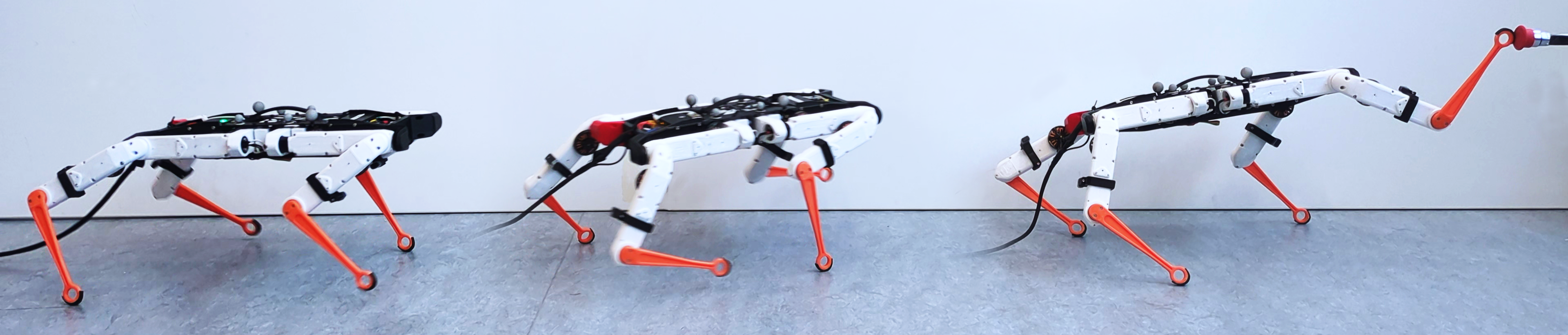} 
    \captionof{figure}{
    Our Solo12 loco-manipulation policy is able to execute and smoothly transition between locomotion and manipulation to reach a specified target.
    } 
\label{fig:solo12_hardware_pedipulation}  
}
\makeatother

\maketitle
\thispagestyle{empty}
\pagestyle{empty}

\begin{abstract}
Although humanoid and quadruped robots provide a wide range of capabilities, current control methods, such as Deep Reinforcement Learning, focus mainly on single skills. This approach is inefficient for solving more complicated tasks where high-level goals, physical robot limitations and desired motion style might all need to be taken into account. A more effective approach is to first train a multipurpose motion policy that acquires low-level skills through imitation, while providing latent space control over skill execution. Then, this policy can be used to efficiently solve downstream tasks. This method has already been successful for controlling characters in computer graphics. In this work, we apply the approach to humanoid and quadrupedal loco-manipulation by imitating either simple synthetic motions or kinematically retargeted dog motions. We extend the original formulation to handle constraints, ensuring deployment safety, and use a diffusion discriminator for better imitation quality. We verify our methods by performing loco-manipulation in simulation for the H1 humanoid and Solo12 quadruped, as well as deploying policies on Solo12 hardware. Videos and code are available at \url{https://gepetto.github.io/LaCoLoco/}
\end{abstract}

\section{Introduction}

Traditional methods of robot control such as reinforcement learning and trajectory optimization have been successful in enabling legged robots to perform a wide variety of skills such as walking, running or even parkour \cite{rl_review}. These methods work by optimizing a cost metric that encodes information about task success, motion style, constraints and other characteristics of the desired robot motion.

Despite the power and generality of such techniques, designing this metric is a laborious process, commonly known to require extensive trial and error. For example, while defining the cost necessary to achieve single motions is feasible, balancing individual costs to optimize a multipurpose policy is not trivial. Qualities such as the style that motions should follow can require multiple cost terms that need to be carefully designed, like in \cite{learning_to_walk} \cite{bipedal_rl_cassi}.

Imitation learning is a promising approach that removes the need for explicit cost functions. Instead, policies are optimized in a data-driven way, by aiming to generate motions that imitate demonstration examples. Instead of encoding a desired behavior through cost functions, one only needs to provide examples to be imitated.

In computer graphics, controlling physically-simulated characters poses a similar problem to robot control. In order to achieve reusable character control policies, \cite{ase} introduced Adversarial Skill Embeddings (ASE), a method for imitating motion datasets and generating actions via latent space control. The generative approach used in ASE, enables combining and transitioning smoothly between learned motions. Furthermore, it is possible to simultaneously imitate different motion datasets, regardless of their source. By observing only states instead of state-action pairs \cite{gail} \cite{drail} the approach can also be applied to sources such as animal motion demonstrations, where the action space is not observable \cite{amp_hardware}. 

Motion generation that is controllable through a meaningful latent space instead of working directly in configuration space can be beneficial when solving complex control tasks. A latent space control policy can be interpreted as a decoder from latent to configuration space that ensures the generated motions will be close to the motions it was trained to imitate.

We seek to better utilize the capabilities of legged robots by developing policies that can perform both walking and manipulation skills. In this work, we explore the application of ASE to humanoid and quadrupedal loco-manipulation by first training low-level imitation policies on various motion datasets. Then, we use these policies to solve the problem of reaching points with an end-effector. To facilitate safe hardware deployment, we extend the original method to allow for explicit constraints during training. Additionally, we propose an alternative discriminator to achieve better quality imitation. \\

\noindent This work proposes the following contributions:

\begin{itemize}
    \item Development of multi-skill latent conditioned policies for humanoid and quadrupedal loco-manipulation through adaptation of ASE
    \item Demonstration of successful imitation of synthetic motions from different base controllers, as well as trajectories obtained from dog motion capture data
    \item Extension of the original ASE method with stochastic constraints and diffusion-based discriminator 
    \item Validation and comparison of methods on the task of end-effector point tracking for simulated H1 and Solo12 robots as well as Solo12 hardware
\end{itemize}

To the best of our knowledge, this work is the first to successfully apply latent conditioned imitation learning and demonstrate autonomous skill switching between locomotion and manipulation on a quadruped hardware platform.

\section{Related work}

\subsection{Motion Imitation}

Imitation learning is a data-driven approach for designing controllers by aiming to mimic a dataset of expert behaviors. GAIL \cite{gail} introduced an imitation learning technique for physics-based control tasks based on Generative Adversarial Networks (GANs) \cite{gan} and reinforcement learning. This approach was further extended by AMP \cite{amp} to train stylized task-specific policies, by formulating a style reward as imitation. Subsequently, \cite{amp_hardware} applied AMP to the task of quadrupedal velocity tracking while also imitating a dog. While this method successfully trained hardware controllers that achieve a task in a desired style, these controllers are single-purpose and cannot be reused to achieve a different task than the one originally trained for.

\subsection{Latent Space Control}

To train task-agnostic policies that can perform a diverse set of skills, \cite{diayn} proposed conditioning a policy on a latent variable and optimizing a mutual information objective. This approach was adapted by \cite{cassi} to learn a multi-skill quadrupedal locomotion policy conditioned on a discrete latent variable, able to perform different types of gaits. This method is not fully multipurpose, as the controller is trained with a task-specific reward. In ASE \cite{ase}, a hierarchical control approach with a reusable low-level controller is trained instead. The controller is conditioned on a continuous latent representation, enabling smooth transitions between skills.

\subsection{Loco-manipulation}

The combination of self-mobility (locomotion) and object interaction (manipulation) has been studied for both humanoid robots \cite{gu2025humanoidlocomotionmanipulationcurrent} and quadrupeds. For quadrupeds, \cite{pedipulation_policy_switching_bt} proposed a dual-policy approach to perform loco-manipulation using one of the robot's legs. A single-purpose controller was proposed instead by \cite{pedipulate}, able to perform both behaviors by optimizing a task specific cost. \cite{pedipulation_diffusion} demonstrated the ability to use a similar low-level controller combined with a high-level diffusion policy to execute tasks such as lifting a basket.

For humanoid robots, loco-manipulation capabilities are crucial for enabling human-like environment interaction. To this end, \cite{hover} proposed a generalist policy for controlling humanoids that is able to track combinations of kinematic, joint or base-related targets. While this scheme is general enough to allow for a variety of behaviors, we speculate that a meaningful latent space would be a more suitable interface for achiving high-level tasks using a general policy.

\subsection{Constraints}

One key difference between animated character control and robotics is that during robot control, it is of vital importance to take physical limitations of the hardware into account. Limits such as maximum torque or kinematics can be modeled as simulation constraints during policy training, however others such as ground reaction forces cannot. In order to explicitly allow for constraints during reinforcement learning, \cite{cat} stochastically terminates future policy rewards on constraint violation, therefore encouraging constraint satisfaction. 

\section{Preliminaries}

Our proposed hierarchical control approach is based on Adversarial Skill Embeddings (ASE) \cite{ase}, with two key differences.
Firstly, our method supports constraints through stochastic reward termination \cite{cat}, without requiring difficult-to-tune reward penalties. Taking constraints into account enables us to train safer and better-behaved policies for hardware deployment.
Secondly, we use a diffusion discriminator instead of the original GAN-style model to achieve higher quality imitation.

\subsection{Adversarial Skill Embeddings}

The ASE framework \cite{ase} consists of two stages: a \textit{general-purpose} low-level policy controllable by a continuous latent $z$, and a high-level \textit{task-specific} policy that generates latents to achieve a desired goal.

The low-level policy $\pi(a|s,z)$ provides a distribution over agent actions $a$ that would result in motions imitating a provided dataset $M$. The generated behavior depends on information encoded by the latent \textit{skill variable} $z$. The policy $\pi$ is trained using reinforcement learning, based on the standard discounted Markov Decision Process (MDP) framework:
\begin{equation} \label{mdp}
\begin{aligned}
    \max_{\pi} \quad \mathbb{E}_{p(\tau | \pi)} 
    \left[ \sum_{t=0}^{T-1} \gamma^t r_t \right], 
\end{aligned}
\end{equation}
where $\gamma$ is the discount factor, $T$ the time horizon, $\tau=\{ s_0, a_0, r_0, s_1, ...., s_{T-1}, a_{T-1}, r_{T-1}, s_{T} \}$ a trajectory sampled from the policy $\pi$, and $r_t$ the reward at timestep $t$.

For the low-level policy, a combined reward is optimized:
\begin{equation} \label{low_level_reward}
\begin{aligned}
    r_t = r_D(s_t, s_{t+1}) + \beta r_E(s_t, s_{t+1})
\end{aligned}
\end{equation}
where $r_D$ is a motion imitation objective, $r_E$ a skill discovery objective, and $\beta$ a weight hyperparameter. We provide more details on these objectives below.

After $\pi$ is trained on a motion dataset $M$, it can be used for downstream tasks. In ASE, a high-level policy is trained to generate latents $z$ that guide $\pi$ towards achieving task-specific goals. In the original formulation, an additional \textit{motion prior} reward is used when training the high-level policy which we omit for simplicity. \\

\subsubsection{Imitation objective}
%%%%%%%%%%%%%%%%%%%%%%%%%%%
The low-level policy imitation term $r_D$ encourages the distributions of state transitions from the policy $d^\pi(s_t, s_{t+1})$ and the dataset $d^M(s_t, s_{t+1})$ to match as closely as possible:
\begin{equation} \label{jensenshannon}
    r_D = -JS(d^\pi(s_t, s_{t+1}) \parallel d^M(s_t, s_{t+1}))
\end{equation}
where $JS$ is the Jensen-Shannon distribution divergence. Maximizing $r_D$ means that the policy will generate motions that imitate the dataset. 
%%%%%%%%%%%%%%%%%%%%%%%%

Calculating the $JS$ divergence as written in Eq. \ref{jensenshannon} can be computationally intractable. To combat this issue, ASE \cite{ase} uses a GAN-based variational approximation to calculate the distribution distance, introduced in \cite{gan}:
\begin{equation} \label{gan}
\begin{aligned}
\min_{D} \quad & - \mathbb{E}_{d^M(s_t, s_{t+1})} && \left[ \log(D(s_t, s_{t+1})) \right] \\\
    &- \mathbb{E}_{d^{\pi}(s_t, s_{t+1})} && \left[ \log (1 - D(s_t, s_{t+1})) \right]
\end{aligned}
\end{equation}
where $D(\cdot, \cdot)$ is a \textit{discriminator} function that aims to classify whether a state transition was produced by the policy or is part of the dataset.

In ASE, $D(s_t, s_{t+1})$ is calculated directly by a neural network $D_{net}$.

The discriminator network $D_{net}$ and the low-level policy are trained at the same time. During training, the discriminator aims for the generated behaviors to be (from the discriminator's perspective) as if they originated from the motion dataset.

The policy reward term is:
\begin{equation}
    r_D = -log(1 - D_{net}(s_t, s_{t+1}))
\end{equation}

\subsubsection{Skill Discovery Objective}

%%%%%%%%%%%%%%%%%%%%%%%%
Aside from the imitation objective $r_D$, ASE additionally uses a \textit{skill discovery} term $r_E$ while training the low-level policy. This term maximizes the mutual information $I(\cdot, \cdot)$ between $z$ and state transitions $(s_t, s_{t+1})$ produced by $\pi$:

\begin{equation}
\begin{aligned}
    r_E(s_t, s_{t+1}) &= I(s_t, s_{t+1}; z | \pi) \\
    &= \mathcal{H}(s_t, s_{t+1} | \pi) - \mathcal{H}(s_t, s_{t+1} | z, \pi)
\end{aligned}
\end{equation}
where $\mathcal{H}(\cdot)$ is the entropy function.

As in \cite{diayn}, optimizing $r_E$ ensures that the distribution of state transitions is specific to that $z$; in other words, $z$ will control the behavior generated by the low-level policy. However, due to the fact that optimizing this formulation is intractable, the following lower bound is maximized instead:

\begin{equation} \label{mutual_info}
\begin{aligned}
    I(s_t, s_{t+1}; z | \pi) \geq &\max_{q} \mathcal{H}(z) \\
    &+ \mathbb{E}_{p(z)} \mathbb{E}_{p(s_t, s_{t+1} | \pi, z)} \left[ \log q(z | s_t, s_{t+1}) \right]
\end{aligned}
\end{equation}

where $q(z|s_t, s_{t+1})$ is called the skill encoder. The encoder $q(z|s_t, s_{t+1})$ is modeled as a von Mises-Fisher distribution:
\begin{equation}
\begin{aligned}
    q(z | s_t, s_{t+1}) = \frac{1}{Z}exp(\kappa \mu_q(s_t, s_{t+1})^Tz)
\end{aligned}
\end{equation}
where $\mu_q$ is predicted by a neural network, $\kappa$ is a scaling factor and $Z$ a normalization constant. In order to achieve the maximization in Eq. \ref{mutual_info}, the encoder is trained jointly during policy training. The expectations are approximated by using samples of $z$ and state transitions:

\begin{equation}
    \max_{q} \quad \mathbb{E}_{p(z)} \mathbb{E}_{d^{\pi}(s_t, s_{t+1}|z)} 
    \left[ \kappa \, \mu_q(s_t, s_{t+1})^{T} z \right]
\end{equation}

In this case, the distribution of $z$ is assumed to be constant, so $H(z)$ does not change and can be omitted. Using the above, the low-level policy \textit{skill discovery} reward that maximizes $I(s_t, s_{t+1}; z|\pi)$ is therefore:

\begin{equation}
\begin{aligned}
    r_E(s_t, s_{t+1}) = \log q(z_t | s_t, s_{t+1})
\end{aligned}
\end{equation}

\subsubsection{Diversity Term}

In addition to optimizing the low-level policy reward, ASE uses a separate \textit{diversity} objective during optimization that enourages similar latents $z$ to produce similar actions:
\begin{equation}
\begin{aligned}
    \mathbb{E}_{d^{\pi}(s)} \mathbb{E}_{z_1, z_2 \sim p(z)} 
    \left[ \left( \frac{D_{\text{KL}} \left( \pi(\cdot | s, z_1), \pi(\cdot | s, z_2) \right)}
    {D_z(z_1, z_2)} - 1 \right)^2 \right]
\end{aligned}
\end{equation}
where $D_z(z_1, z_2) = 0.5(1-z_1^Tz_2)$ is the cosine distance between latents and $D_{KL}(\cdot, \cdot)$ the KL divergence.

\subsubsection{Latent Space}

The high-level policy in ASE generates unnormalized latents that are then projected to the hypersphere before being passed to the low-level controller. This is so that the high-level policy is able to control the entropy of the distribution of $z$, allowing for skill specialization.

During training of the low-level policy, $z$ is drawn from a latent space defined as the hypersphere with unit norm:
\begin{equation}
\begin{aligned}
    \bar{z} \sim \mathcal{N}(0,I),\quad z=\frac{\bar{z}}{||\bar{z}||}
\end{aligned}
\end{equation}

\subsection{Constraints as Terminations}
Constraints as Terminations (CaT) \cite{cat} introduced a method for enforcing explicit constraints during MDP optimization by extending the original formulation in Eq. \ref{mdp}. CaT encourages constraint satisfaction by including a probability of future reward termination depending on whether the policy violates the constraints:
\begin{equation}
\begin{aligned}
    \max_{\pi} \quad \mathbb{E}_{p(\tau | \pi)} 
    \left[ \sum_{t=0}^{T-1} \left( \prod_{t'=0}^{t} \gamma \left( 1-\delta(s_{t'}, a_{t'}) \right) \right) r_t \right]
\end{aligned}
\end{equation}
where $\delta(s, a)$ is a random variable defined as a function of constraint violation $c_i(s,a)$. The $i$-th constraint is violated at state $s$ with action $a$, if $c_i(s, a) > 0$.

This formulation probabilistically terminates policy reward if constraints are violated, steering the optimization towards obeying the constraints.

CaT uses an intuitive $\delta$ which increases when constraints are violated more, up to a maximum probability $p_i^{max}$:
\begin{equation}
\begin{aligned}
    \delta = \max_{i \in I} p_{i}^{max} \cdot \text{clip} \left( \frac{c_{i}^{+}}{c_{i}^{max}}, 0, 1 \right)
\end{aligned}
\end{equation}
where $c_{i}^{+}=\max(0, c_i(s,a))$ is the violation of constraint $i$, and $c_{i}^{max}$ a filtered measure of maximum constraint violation during the last batch of interaction with the environment.

\subsection{Diffusion-Reward Adversarial Imitation Learning}
An alternative to imitation learning approaches such as \cite{gail}, which use a discriminator neural network trained by Eq. \ref{gan}, is using diffusion models \cite{ddpm} as a discriminator. This was used and extended by Diffusion-Reward Adversarial Imitation Learning (DRAIL) \cite{drail}. The enhanced conditional diffusion discriminator of DRAIL was shown to produce a more robust reward value and exhibit less overfitting to expert demonstrations than the GAN-based method.

Instead of the original state-action formulation, we modify DRAIL slightly to work on state transitions. The DRAIL discriminator aims to classify whether a transition was part of an expert demonstration or generated by the policy. This is achieved in a single denoising step, by calculating the \textit{diffusion loss}:

\begin{equation} \label{diff_loss}
\mathcal{L}_{\text{diff}}(s_t, s_{t+1}, l) = \mathbb{E}_{t_d \sim T_d} \left[ \left\| \epsilon_{\phi}(s_t, s_{t+1}, \epsilon, t_d | l) - \epsilon \right\|^2 \right]
\end{equation}
where $(s_t, s_{t+1})$ is a state transition, $\epsilon \sim \mathcal{N}(0,I)$ the applied noise, 
$\epsilon_{\phi}$ the predicted noise, $t_d$ the diffusion timestep and $l \in \{l^M, l^\pi\}$ a binary condition label where $l^M$ denotes a transition from the dataset and $l^\pi$ one from the policy.

$\mathcal{L}_{\text{diff}}$ is used to measure how well the distribution of policy state transitions matches the motion dataset by comparing noise predictions. For example, if a state transition was sampled from the dataset, the predicted noise for label $l^M$ would match the actual noise, setting $\mathcal{L}_{\text{diff}}(s_t, s_{t+1}, l^M)$ close to 0. $\mathcal{L}_{\text{diff}}(s_t, s_{t+1}, l^\pi)$ would in this case be large. The reverse applies to transitions generated by the policy.

In order to obtain a reward signal from $\mathcal{L}_{\text{diff}}(s_t, s_{t+1}, l)$, a sigmoid classifier is constructed:
\begin{equation}
\begin{aligned}
    D_{\text{diffusion}}(s_t, s_{t+1}) = \sigma \big(& \mathcal{L}_{\text{diff}}(s_t, s_{t+1}, l^\pi)\\
    &-\mathcal{L}_{\text{diff}}(s_t, s_{t+1}, l^M) \big)
\end{aligned}
\end{equation}
where $\sigma(\cdot)$ is the sigmoid function. As in the GAN-style formulation \cite{gail}, the low-level policy reward is defined as:
\begin{equation}
\begin{aligned}
    r_D(s_t, s_{t+1}) = - \log \left( 1 - D_{\text{diffusion}}(s_t, s_{t+1}) \right)
\end{aligned}
\end{equation}

where $D(s_t, s_{t+1})$ is provided in this case by the diffusion model. To calculate the expectation in Eq. \ref{diff_loss}, multiple $t_d$ are sampled during training. The $\epsilon_{\phi}$ network is trained directly using the discriminator loss.

\section{Method}
By leveraging unstructured motion datasets, our method aims to learn reusable robot policies that can perform and switch between multiple skills. We base our work on the hierarchical framework of Adversarial Skill Embeddings (ASE) with two key modifications:

\begin{itemize}
    \item We use the diffusion discriminator introduced by DRAIL \cite{drail} instead of the GAN-based discriminator in ASE \cite{ase}. We have qualitatively observed that the diffusion discriminator generally results in higher-quality motions that are closer to the demonstration dataset. \footnote{Please refer to the supplementary video for a comparison of motions generated by each method.}
    
    \item We allow for constraints to be explicitly specified during low-level policy training by using the Constraints as Terminations (CaT) \cite{cat} method. This is crucial for successful deployment of our method to hardware, as the policy can be trained to respect limits which are difficult to enforce during simulation. We show that limiting ground reaction forces during imitation results in more gentle policies and smoother motions. We also demonstrate that enforcing constraints on the low-level policy is enough; motions generated by a task-specific high-level policy still respect constraints set during low-level policy training. \\

\end{itemize}

These modifications proved key in enabling our method to generate high-quality, safe, multi-skill policies that are applicable to robot hardware.

\begin{figure}[t]
  \vspace{1em}
  \centering
  \includegraphics[width=0.3\textwidth]{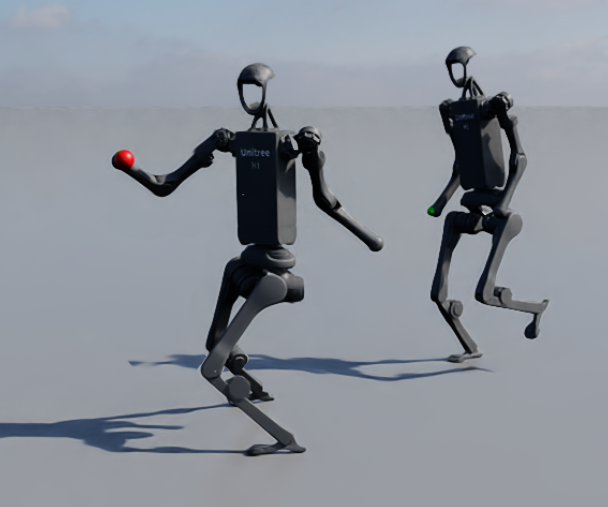}
  \caption{H1 policy transitioning from walking to reaching}
\label{h1_switch_fig}
\end{figure}

\section{Application to Humanoid and Quadrupedal Loco-manipulation}

We demonstrate the effectiveness of our method in imitating a variety of motion datasets and learning reusable skills, by focusing on humanoid and quadrupedal loco-manipulation. We select the loco-manipulation task in order to investigate whether a policy trained with our method can learn to perform and switch smoothly between the distinct motions required when locomoting and reaching for a point in space. Furthermore, we would like to investigate whether the latent space learned by training the low-level policy allows for precise robot control. 

We formulate the loco-manipulation task similarly to \cite{pedipulate}. The task reward to be optimized is a function of end-effector distance from a point in 3D space:

\begin{equation} \label{locomanip_reward}
\begin{aligned}
    r_g = exp(-10 ||x_{eef} - x_{d}||)
\end{aligned}
\end{equation}

where $x_{eef}$ is the current end-effector position and $x_{d}$ the position of the destination point.

For our humanoid experiments, we test using the Unitree H1 robot in simulation and select the right hand as the end-effector. For our quadruped experiments, we use the Solo12 open-source robot \cite{odri} in simulation and real life. The front right foot is selected as the end-effector. 

\section{Experimental Setup}
To evaluate the behavior of our method, examine the generated motions and verify its applicability to hardware, we train policies for the loco-manipulation task on various motion datasets. For H1, we compile a synthetic dataset of separate walking and reaching-in-place motions. These correspond to the two fundamental skills required for loco-manipulation. We also generate a similar dataset for Solo12.

To further demonstrate the generality of our method, we investigate whether the low-level policy can be also trained to imitate physically-infeasible motions. We use a dataset of dog motions collected using a motion capture system \cite{dog_mocap}, that we kinematically retarget to the morphology of Solo12 without regard to dynamic feasibility.   

For all experiments, we perform an ablation comparing the base ASE \cite{ase} method, ASE with the DRAIL \cite{drail} discriminator, as well as both versions trained using CaT \cite{cat} to limit ground reaction forces. We examine the end-effector tracking accuracy of each method, the amount of time that constraints are in violation and how many times the robot fell while executing the task.

\subsection{Motion Datasets}

\subsubsection{Synthetic Datasets}

For both H1 and Solo12, we generate separate locomotion and reach-in-place motions for low-level policy training. For H1, we obtain walking trajectories by training the example velocity tracking policy in IsaacLab \cite{isaaclab} and randomly sampling velocity commands. We obtain reaching trajectories by randomly sampling close position commands and training a simple policy that optimizes end-effector position tracking. We follow a similar procedure for the Solo12 dataset. We generate walking motions using the policy in \cite{cat} and reaching motions using the pedipulation policy in \cite{pedipulate}.

In both cases, we sample targets for the reaching motions that are close enough to the robot to be reachable without walking. Overall, we collect 120 examples of walking and 20 examples of reaching, each lasting 10 seconds. \\

\subsubsection{Dog Motion Dataset}

\begin{figure}[t]
  \vspace{1em}
  \centering
  \includegraphics[width=\columnwidth]{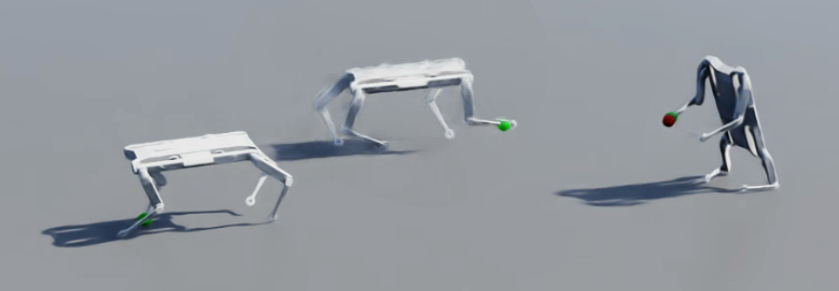}

  \caption{Solo12 policy learns to extract and switch between skills encountered in dog motion dataset}
    \label{solo_switch_fig}
\end{figure}

We generate a dataset of dog motions for the Solo12 by kinematically retargeting motion capture (mocap) data from \cite{dog_mocap}. For each trajectory in the source mocap data, we formulate an optimization problem per timestep to find a robot configuration that minimizes the distance of selected robot frames from the equivalent dog keypoints:
\begin{equation}
    \min_{q_t} \sum_{k=1}^K{\norm{p_{fk}^k(q_t) - \hat{p}^{k}_t}^2} + \alpha \norm{q_t^{joints} - q_{neutral}^{joints}}^2
\end{equation}

where $q_t$ is the configuration of the robot at timestep $t$, $K$ the number of keypoints to be matched, $p_{fk}^k(q_t)$ the world position of the $k$-th frame at configuration $q_t$, $\hat{p}_t^k$ the equivalent position of frame $k$ at time $t$ in the mocap motion and $\alpha=0.005$ a weight factor. The second regularization term encourages the joint positions $q_t^{joints}$ to not deviate too much from a default joint configuration $q_{neutral}^{joints}$. 

We calculate $p_{fk}^k(q_t)$ for each frame using forward kinematics. For our experiments, we select the foot and shoulder frames to be matched by the optimization, eight in total. This is adequate to obtain motions that are close to the source set. We optimize using Ipopt \cite{ipopt} by warm-starting the next timestep with the output configuration of the previous timestep. 

\subsection{Observations}

Drawing inspiration from \cite{amp_hardware}, our discriminator has access to a subset of the full robot state (observation space) to use as features when producing the imitation reward. For both Solo12 and H1, we construct this subset using the following components: base height, base linear and angular velocities, gravity vector projection to the base frame, joint positions and velocities. The observation space used for the low-level policy consists of the same state components, with the addition of the last generated policy action.

\subsection{Training}

We use the Isaac Lab \cite{isaaclab}  simulator to train policies using Proximal Policy Optimization (PPO) \cite{ppo}. For all methods, we simulate 4096 environments in parallel. To facilitate easier transfer of controllers from simulation to hardware, we add noise to policy observations and randomize frictions and masses during training.

\subsubsection{Low-level Policy}

For all methods, the frequency of the low-level policy is 50 Hz. The policy is trained for 7000 epochs and uses a 7-dimensional continuous latent space.

For methods using CaT, we limit the maximum ground reaction force per leg to ensure robot movements are gentle on the hardware. For H1, we set this limit to 1.5 times the weight of the robot. For Solo12 we use $25$ N which is approximately the weight of the robot. To encourage skill acquisition, we set $p_i^{max} = 0$ at the beginning of training. Towards the end, $p_i^{max}$ is increased linearly to 0.1 for H1 and 0.2 for Solo12, to ensure adherence to constraints.

\subsubsection{High-level Policy}

For all experiments, we train the high-level policy using the end-effector tracking cost of Eq. \ref{locomanip_reward} for 2000 epochs. The frequency of the policy is 10 Hz.

To train the Solo12 controller, we sample destination points on a cylinder of radius $1.5$ m at height in the range $(0.16, 0.42)$ m. To ensure that the robot would have to combine both locomotion and reaching to achieve the goal, we randomly initialize its position near the center of the circle.

For H1, we always initialize the robot at the origin. We randomly sample destination points in the rectangle $x: (0.3, 4.0)$, $y: (-1.0, 1.0)$, $z: (1.0, 1.4)$ m.

\section{Results}

In all loco-manipulation experiments, we compute the end-effector position error by using the last episode step.  We train all four methods on each dataset using 5 seeds and report the best policy, picked based on performance and fall count. For the Solo12 hardware experiments we deploy the best controllers as measured in simulation.

\begin{table*}
\centering
\vspace*{1em}
\caption{Simulation results for H1 and Solo12 ($\downarrow$) (best policy among 5 seeds)}

\begin{tabular}{|c|c|c|c|c|c|c|}
\hline
\rule{0pt}{0.9\normalbaselineskip}
\textbf{Experiment} & \textbf{Method} & \textbf{\thead{End-Effector \\Mean Error ± Std [cm]}} & \textbf{\thead{Contact Force\\Violation [\%]}} & \textbf{Falls [\%]} \\
\hhline{-|-|-|-|-|-|-}
\rule{0pt}{0.9\normalbaselineskip}
\multirow{4}{*}{H1 Synthetic} & ASE & 14.43 ± 8.18 & 0.7189 & 0.0244 \\ 
    & ASE DRAIL & 7.46 ± 3.81 & 1.4394 & 0.0000 \\ 
    & ASE CaT & 11.51 ± 6.11 & 0.0198 & 0.3418 \\ 
    & ASE DRAIL CaT & 10.24 ± 5.76 & 0.3249 & 0.0000 \\[0.5mm]
\hline
\rule{0pt}{0.9\normalbaselineskip}
\multirow{4}{*}{Solo12 Synthetic} & ASE & 3.45 ± 1.44 & 0.6632 & 0.0488 \\ 
    & ASE DRAIL & 2.72 ± 3.89 & 0.8725 & 0.1709 \\ 
    & ASE CaT & 3.47 ± 7.35 & 0.0426 & 0.0000 \\ 
    & ASE DRAIL CaT & 2.51 ± 2.74 & 0.0892 & 0.0488 \\ [0.5mm]
\hline
\rule{0pt}{0.9\normalbaselineskip}
\multirow{4}{*}{Solo12 Dog Imitation} & ASE & 19.98 ± 12.18 & 1.6981 & 0.0000 \\ 
    & ASE DRAIL &  9.74 ± 16.36 & 0.7926 & 0.8301 \\ 
    & ASE CaT & 26.64 ± 14.08 & 0.1433 & 0.7812 \\ 
    & ASE DRAIL CaT & 11.85 ± 15.95 & 0.1632 & 0.7568 \\[0.5mm]
\hline
\end{tabular}
\label{sim_results}
\end{table*}

\subsection{Simulation}

In simulation, all methods successfully train policies that perform and switch between skills to achieve the loco-manipulation task, as seen in Figures \ref{h1_switch_fig} and \ref{solo_switch_fig}. Surprisingly, this is also the case for methods imitating the retargeted dog motion dataset.

The reaching performance of all methods that train on the synthetic datasets is similar, as can be seen from Table \ref{sim_results}. In these experiments, methods using the DRAIL discriminator yielded only slightly better errors. In experiments that train on the dog motion dataset, DRAIL results in a more significant improvement.

All methods that use CaT during training to limit ground reaction forces show a reduced constraint violation time. The fall count over 4096 episodes is generally small and does not seem to follow an obvious trend.

Training-wise, all Solo12 policies demonstrate some switching from locomotion to reaching. For H1 however, DRAIL methods are more unstable to train, despite the increased accuracy achieved in the best case. We observed that in most cases, policies trained with DRAIL for H1 did not learn to switch motions for the entire episode.   

Qualitatively, we notice that methods trained on the synthetic datasets generally behave similarly. The difference between methods is significantly more pronounced when training on the dog motion dataset. In this case, the original ASE method tends to generate motions that are more harsh and oscillatory. Adding CaT results in visibly more graceful motions. The DRAIL discriminator has the largest impact on motion quality. DRAIL results in significantly smoother movement that resembles the original dataset better, as well as stable reaching. Please refer to the supplementary video for a visual comparison of these policies.

\subsection{Hardware}

In addition to simulation tests, we use a real-life Solo12 robot to evaluate all four policies trained using the synthetic motion dataset. The results are presented in Table \ref{hardware_solo}. We collect 20 measurements for each method, starting from random positions and orientations. The end-effector target position is set to $\hat{p} = (0, 0, h)$, where $h$ is a random height sampled from the range $(0.16, 0.42)$ m for each trial but shared across all methods. We use a motion capture system to track the position and linear velocity of the robot.

\begin{table}[h]
\centering
\caption{Hardware results for Solo12 Synthetic ($\downarrow$) (best policy in simulation among 5 seeds)}

\begin{tabular}{|c|c|c|c|c|c|c|}
\hline
\textbf{Method} & \textbf{\thead{End-Effector \\Mean Error ± Std [cm]}} & \textbf{Falls [\%]} \\
\hhline{-|-|-}
\rule{0pt}{0.9\normalbaselineskip}
ASE & 7.25 ± 3.37 & 0.00 \\ 
ASE DRAIL & 5.64 ± 5.52 & 5.00 \\ 
ASE CaT & 9.03 ± 6.19 & 0.00 \\ 
ASE DRAIL CaT & 5.62 ± 2.76 & 0.00 \\[0.5mm]
\hline
\end{tabular}
\label{hardware_solo}
\end{table}

All policies are successful in combining locomotion and reaching to achieve the loco-manipulation task. Accuracy is similar for all, with DRAIL achieving only slightly better results.

The qualitative difference in policy behavior however, is more significant. ASE without CaT generates aggressive locomotion that jumps when reaching the target, leading to excessive stress on the hardware. With our Solo12, we observe that these motions are likely to trigger a hardware issue, where motor drivers are sometimes unable to track high accelerations. When this occurs, we repeat the experiment as the robot shuts down.

Methods trained with DRAIL generate more gentle locomotion trajectories on the hardware, which resemble the motion dataset more closely. Furthermore, as expected, policies trained with CaT are generally even less aggressive and more stable. No policy resulted in significant loss of balance, with only ASE DRAIL falling in 1 out of 20 trials.

We did not manage to successfully deploy any of the policies trained on the dog motion dataset. Unfortunately, no policy was able to balance long enough when walking to achieve the task. We hypothesize that the gait style in the dog dataset is more difficult to learn and execute, since the feet are close together when walking. This results in a small support polygon, making balance more demanding. Given that all methods were successful in simulation but not in real life, model mismatch might be a potential explanation why these policies did not work.

\section{Discussion and Future work}

In this work, we demonstrate the application of hierarchical latent conditioned control in robotics. We show that by training on unlabeled motion datasets, a single controller can extract reusable skills that can be used to achieve high-level tasks on hardware. This approach does not require expensive tuning of cost functions, or annotations of the source dataset. 

Our experiments prove that providing a latent space interface to a task-specific controller is adequate to enable precise robot control. We observe that policies select and switch between distinct locomotion and reaching behaviors in order to achieve a task in both simulation and hardware.

In addition, we adapt the original ASE method to use a diffusion discriminator, as well as support constraints for safer hardware deployment. We explore the quantitative and qualitative differences of these methods and demonstrate successful imitation of motions from different sources. We finally show that these methods work on real hardware, and perform ablation experiments to examine the impact of our proposed improvements.  

While our approach is promising, this is still preliminary work. We consider the number of imitated motions and high-level tasks we trained to be limited. Furthermore, the applicability of our H1 policies on hardware is yet to be validated. Other open problems include extending the method to locomotion on rough terrain and integrating perception information to the high-level policy.

\section*{Acknowledgements}

We would like to thank Thomas Flayols and Sabrina Evans for their insightful comments and suggestions.

This research was supported by ANITI IA Cluster project 23-IACL-0002 and ANR-19-PI3A-0004.
This work was granted access to the HPC resources of IDRIS under the allocation 2025-AD011016368 made by GENCI.

\bibliographystyle{IEEEtran}
\bibliography{lacoloco}

\begin{thebibliography}{10}
\providecommand{\url}[1]{#1}
\csname url@rmstyle\endcsname
\providecommand{\newblock}{\relax}
\providecommand{\bibinfo}[2]{#2}
\providecommand\BIBentrySTDinterwordspacing{\spaceskip=0pt\relax}
\providecommand\BIBentryALTinterwordstretchfactor{4}
\providecommand\BIBentryALTinterwordspacing{\spaceskip=\fontdimen2\font plus
\BIBentryALTinterwordstretchfactor\fontdimen3\font minus
  \fontdimen4\font\relax}
\providecommand\BIBforeignlanguage[2]{{%
\expandafter\ifx\csname l@#1\endcsname\relax
\typeout{** WARNING: IEEEtran.bst: No hyphenation pattern has been}%
\typeout{** loaded for the language `#1'. Using the pattern for}%
\typeout{** the default language instead.}%
\else
\language=\csname l@#1\endcsname
\fi
#2}}

\bibitem{rl_review}
C.~Tang, B.~Abbatematteo, J.~Hu, R.~Chandra, R.~Martín-Martín, and P.~Stone,
  ``Deep reinforcement learning for robotics: A survey of real-world
  successes,'' \emph{Annual Review of Control, Robotics, and Autonomous
  Systems}, vol.~8, no. Volume 8, 2025, pp. 153--188, 2025.

\bibitem{learning_to_walk}
N.~Rudin, D.~Hoeller, P.~Reist, and M.~Hutter, ``Learning to walk in minutes
  using massively parallel deep reinforcement learning,'' in \emph{Conference
  on Robot Learning}, 2022.

\bibitem{bipedal_rl_cassi}
Z.~Li, X.~B. Peng, P.~Abbeel, S.~Levine, G.~Berseth, and K.~Sreenath,
  ``Reinforcement learning for versatile, dynamic, and robust bipedal
  locomotion control,'' \emph{The International Journal of Robotics Research},
  vol.~44, no.~5, pp. 840--888, 2025.

\bibitem{ase}
X.~B. Peng, Y.~Guo, L.~Halper, S.~Levine, and S.~Fidler, ``Ase: large-scale
  reusable adversarial skill embeddings for physically simulated characters,''
  \emph{ACM Trans. Graph.}, vol.~41, no.~4, July 2022.

\bibitem{gail}
J.~Ho and S.~Ermon, ``Generative adversarial imitation learning,'' in
  \emph{Proceedings of the 30th International Conference on Neural Information
  Processing Systems}, ser. NIPS'16, 2016, p. 4572–4580.

\bibitem{drail}
C.-M. Lai, H.-C. Wang, P.-C. Hsieh, Y.-C.~F. Wang, M.-H. Chen, and S.-H. Sun,
  ``Diffusion-reward adversarial imitation learning,'' in \emph{Advances in
  Neural Information Processing Systems}, vol.~37, 2024, pp. 95\,456--95\,487.

\bibitem{amp_hardware}
A.~Escontrela, X.~B. Peng, W.~Yu, T.~Zhang, A.~Iscen, K.~Goldberg, and
  P.~Abbeel, ``Adversarial motion priors make good substitutes for complex
  reward functions,'' in \emph{2022 IEEE/RSJ International Conference on
  Intelligent Robots and Systems (IROS)}, 2022, pp. 25--32.

\bibitem{gan}
I.~Goodfellow, J.~Pouget-Abadie, M.~Mirza, B.~Xu, D.~Warde-Farley, S.~Ozair,
  A.~Courville, and Y.~Bengio, ``Generative adversarial networks,''
  \emph{Commun. ACM}, vol.~63, no.~11, p. 139–144, Oct. 2020.

\bibitem{amp}
X.~B. Peng, Z.~Ma, P.~Abbeel, S.~Levine, and A.~Kanazawa, ``Amp: adversarial
  motion priors for stylized physics-based character control,'' \emph{ACM
  Trans. Graph.}, vol.~40, no.~4, July 2021.

\bibitem{diayn}
B.~Eysenbach, A.~Gupta, J.~Ibarz, and S.~Levine, ``Diversity is all you need:
  Learning skills without a reward function,'' \emph{International Conference
  on Learning Representations}, 2019.

\bibitem{cassi}
C.~Li, S.~Blaes, P.~Kolev, M.~Vlastelica, J.~Frey, and G.~Martius, ``Versatile
  skill control via self-supervised adversarial imitation of unlabeled mixed
  motions,'' in \emph{2023 IEEE International Conference on Robotics and
  Automation (ICRA)}, 2023, pp. 2944--2950.

\bibitem{gu2025humanoidlocomotionmanipulationcurrent}
Z.~Gu, J.~Li, W.~Shen, W.~Yu, Z.~Xie, S.~McCrory, X.~Cheng, A.~Shamsah,
  R.~Griffin, C.~K. Liu, A.~Kheddar, X.~B. Peng, Y.~Zhu, G.~Shi, Q.~Nguyen,
  G.~Cheng, H.~Gao, and Y.~Zhao, ``Humanoid locomotion and manipulation:
  Current progress and challenges in control, planning, and learning,''
  \emph{arXiv preprint arXiv:2501.02116}, 2025.

\bibitem{pedipulation_policy_switching_bt}
X.~Cheng, A.~Kumar, and D.~Pathak, ``Legs as manipulator: Pushing quadrupedal
  agility beyond locomotion,'' in \emph{2023 IEEE International Conference on
  Robotics and Automation (ICRA)}, 2023, pp. 5106--5112.

\bibitem{pedipulate}
P.~Arm, M.~Mittal, H.~Kolvenbach, and M.~Hutter, ``Pedipulate: Enabling
  manipulation skills using a quadruped robot’s leg,'' \emph{2024 IEEE
  International Conference on Robotics and Automation (ICRA)}, pp. 5717--5723,
  2024.

\bibitem{pedipulation_diffusion}
Z.~He, K.~Lei, Y.~Ze, K.~Sreenath, Z.~Li, and H.~Xu, ``Learning visual
  quadrupedal loco-manipulation from demonstrations,'' in \emph{2024 IEEE/RSJ
  International Conference on Intelligent Robots and Systems (IROS)}, 2024, pp.
  9102--9109.

\bibitem{hover}
T.~He, W.~Xiao, T.~Lin, Z.~Luo, Z.~Xu, Z.~Jiang, C.~Liu, G.~Shi, X.~Wang,
  L.~Fan, and Y.~Zhu, ``Hover: Versatile neural whole-body controller for
  humanoid robots,'' \emph{arXiv preprint arXiv:2410.21229}, 2024.

\bibitem{cat}
E.~Chane-Sane, P.-A. Leziart, T.~Flayols, O.~Stasse, P.~Souères, and
  N.~Mansard, ``Cat: Constraints as terminations for legged locomotion
  reinforcement learning,'' in \emph{2024 IEEE/RSJ International Conference on
  Intelligent Robots and Systems (IROS)}, 2024, pp. 13\,303--13\,310.

\bibitem{ddpm}
J.~Ho, A.~Jain, and P.~Abbeel, ``Denoising diffusion probabilistic models,'' in
  \emph{Proceedings of the 34th International Conference on Neural Information
  Processing Systems}, ser. NIPS '20, 2020.

\bibitem{odri}
F.~{Grimminger}, A.~{Meduri}, M.~{Khadiv}, J.~{Viereck}, M.~{Wüthrich},
  M.~{Naveau}, V.~{Berenz}, S.~{Heim}, F.~{Widmaier}, T.~{Flayols}, J.~{Fiene},
  A.~{Badri-Spröwitz}, and L.~{Righetti}, ``An open torque-controlled modular
  robot architecture for legged locomotion research,'' \emph{IEEE Robotics and
  Automation Letters}, vol.~5, no.~2, pp. 3650--3657, 2020.

\bibitem{dog_mocap}
H.~Zhang, S.~Starke, T.~Komura, and J.~Saito, ``Mode-adaptive neural networks
  for quadruped motion control,'' \emph{ACM Trans. Graph.}, vol.~37, no.~4,
  July 2018.

\bibitem{isaaclab}
M.~Mittal, C.~Yu, Q.~Yu, J.~Liu, N.~Rudin, D.~Hoeller, J.~L. Yuan, R.~Singh,
  Y.~Guo, H.~Mazhar, A.~Mandlekar, B.~Babich, G.~State, M.~Hutter, and A.~Garg,
  ``Orbit: A unified simulation framework for interactive robot learning
  environments,'' \emph{IEEE Robotics and Automation Letters}, vol.~8, no.~6,
  pp. 3740--3747, 2023.

\bibitem{ipopt}
A.~W{\"a}chter and L.~T. Biegler, ``On the implementation of an interior-point
  filter line-search algorithm for large-scale nonlinear programming,''
  \emph{Mathematical Programming}, vol. 106, pp. 25--57, 2006.

\bibitem{ppo}
J.~Schulman, F.~Wolski, P.~Dhariwal, A.~Radford, and O.~Klimov, ``Proximal
  policy optimization algorithms,'' \emph{arXiv preprint arXiv:1707.06347}, 07
  2017.

\end{thebibliography}

\end{document}